\documentclass[letterpaper, 10 pt, conference]{ieeeconf}  

\IEEEoverridecommandlockouts                              

\usepackage[dvipdfmx]{graphicx} 
\usepackage{amsmath} 
\usepackage{amssymb}  
\usepackage{mathtools}
\usepackage{cite}
\usepackage{color}
\usepackage{array}

\def\reffig#1{{Fig. }\ref{#1}}
\def\reftab#1{{Table }\ref{#1}}
\def\refeqn#1{{Eq. }(\ref{#1})}

\def\red#1{#1}
\def\redbf#1{#1}
\def\bluebf#1{#1}

\title{\LARGE \bf
Incipient Slip Detection by Vibration Injection into Soft Sensor
}

\author{Naoto Komeno and Takamitsu Matsubara
\thanks{*This work was supported by JSPS KAKENHI Grant Numbers JP19H01124 and JP22KJ2288.}
\thanks{Both authors are with the Division of Information Science, Graduate School of Science and Technology, Nara Institute of Science and Technology, Japan:
        {\tt\small komeno.naoto.km6@is.naist.jp, takam-m@is.naist.jp}}%
}

\begin{document}

\maketitle
\thispagestyle{empty}
\pagestyle{empty}

\begin{abstract}
In robotic manipulation, preventing objects from slipping and establishing a secure grip on them is critical.
Successful manipulation requires tactile sensors that detect the microscopic incipient slip phenomenon at the contact surface.
Unfortunately, the tiny signals generated by incipient slip are quickly buried by environmental noise, and precise stress-distribution measurement requires an extensive optical system and integrated circuits.
In this study, we focus on the macroscopic deformation of the entire fingertip's soft structure instead of directly observing the contact surface and its role as a vibration medium for sensing.
The proposed method compresses the stick ratio's information into a one-dimensional pressure signal using the change in the propagation characteristics by vibration injection into the soft structure, which magnifies the microscopic incipient slip phenomena into the entire deformation.
This mechanism allows a tactile sensor to use just a single vibration sensor.
In the implemented system, a biomimetic tactile sensor is vibrated using a white signal from a PZT motor and utilizes frequency spectrum change of the propagated vibration as features.
We investigated the proposed method's effectiveness on stick-ratio estimation and \red{stick-ratio stabilization} control during incipient slip.
Our estimation error and the control performance results significantly outperformed the conventional methods.
\end{abstract}

\section{Introduction}
Humans can stabilize the gripping of objects and perform complex manipulations with minimum effort, regardless of the unknown friction coefficient and the object's weight \cite{Johansson}.
Humans prevent gross slip (i.e., object falling) by perceiving an incipient slip \cite{review_slip}, which is a precursor to the object slip that occurs gradually from the periphery of the contact surface.
This phenomenon occurs when soft skin tissues make contact with a surface, which can be divided into two regions: the stick and the slip.
\red{
The degree of this phenomenon is indicated by the stick ratio, which is the percentage of stick area on the contact surface.
}
Even though the only contact region is confined to a tiny space at the fingertip, humans can accurately perceive slip phenomena.
This ability is enabled with high temporal and spatial resolution by the mechanoreceptors that are naturally present in the skin.
For robots to replace human labor for handling complex manipulation tasks, a high-performance tactile sensor is required that can detect incipient slip and stabilize gripping \cite{review_dexterous}.
For detecting incipient slip, such mechanical features on the contact surface as vibrotactile signal \cite{Dong,Yamada2001,Su,microvibration} and pressure distribution \cite{sony,Ikeda,leverage} are utilized.

\begin{figure}
  \centering
  \includegraphics[width=0.75\hsize]{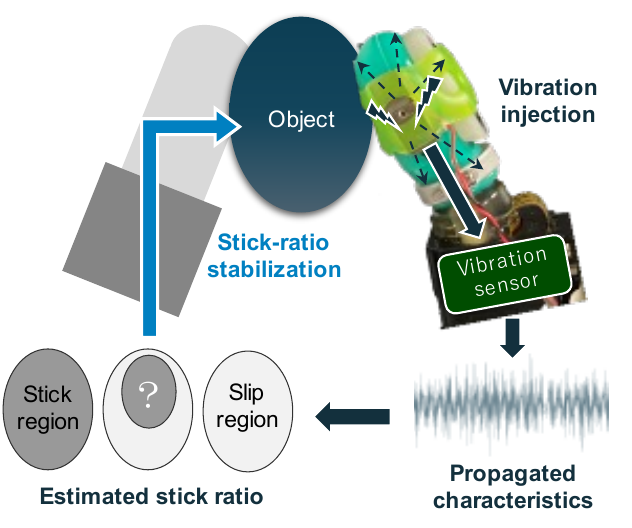}
  \vspace{-2mm}
  \caption{Incipient slip detection with vibration injection achieves \red{stick-ratio stabilization}. Propagated vibration reveals stick ratio and provides a degree of incipient slip and stabilization controller feedback on the information.}
  \label{fig: senzai}
\end{figure}

\red{
Vibrotactile signals, indicating dynamic pressure changes (e.g., pixel displacement, force derivatives, and frequency spectrum), allow unrestricted sensor placement and only need a single pressure sensor for slip detection.
However, their signal-to-noise ratio deteriorates due to low variability at minor friction and grip forces, and they require filters against other signals or noise \cite{review_slip}.
Conversely, pressure distribution produces spatial stress on contact surfaces, shows the extent of slip with stick-ratio calculation, and aids in designing robust controllers.
However, it requires highly integrated sensor arrays and optics, making it less suitable for endurance from mechanical load and dexterous end-effector motion control \cite{active}.
}

Both approaches directly capture the microscopic slip phenomena on rigid or elastic contact surfaces.
Thus, the focus should concentrate on increasing the density of the sensor information within a limited planar space.
Unfortunately, unlike mechanoreceptors in the skin, the integration and wiring of sensor bases have manufacturing costs and engineering limitations, making it impractical to limitlessly improve sensor performance.

By focusing on the soft structure of the biological tissue in the fingertip or the artificial skin material in the tactile sensor, it is possible to avoid completing the microscopic mechanical phenomenon caused by contact and slip only on the contact surface.
The stresses generated at the contact surface are transmitted through the soft structure to the entire finger structure and cause deformation.
Pressure sensors and pneumatic soft robots utilize sensing methods that leverage the property of converting such microscopic signals as contact points and local deformation shapes into macroscopic signals by the soft structure and manifest as sound or acceleration throughout the structure \cite{softrobot,softfinger}.
Hence, a similar mechanism could be utilized for incipient slip detection rather than relying on the microscopic phenomena in contact surfaces.
We propose a tactile sensor with a soft structure that can be used as an amplifier to magnify the change in the stick ratio due to microscopic incipient slip.

This study proposes a new incipient slip detection method that uses a vibration injection into the soft structure of a tactile sensor, which is macroscopically deformed with stick ratio's changes (\reffig{fig: senzai}).
The vibration propagates through the soft structure, and the characteristics that contain information of the deformation allow the stick ratio to be estimated.
In our experimental setup, a PZT motor provides a white signal to a biomimetic tactile sensor with a sufficiently soft structure to act as a vibration medium.
The tactile sensor's internal vibration sensor measures the frequency spectrum of the vibration injection.
We conducted experiments to verify the \red{stick-ratio stabilization} control performance using real-time incipient slip detection with a regressed slip model for estimating the stick ratio.
This investigation allowed us to compare the effectiveness of the proposed method with the conventional vibrotactile signal and pressure distribution methods.
The performance of the real-time \red{stick-ratio stabilization} control was evaluated as the stick ratio decreased.

\section{Related works}

\subsection{Incipient slip detection}
\subsubsection{Vibrotactile signal}
Dong et al. proposed a method to detect the slip from the elastomer deformation detected by the pixel displacement of surface markers using a GelSight sensor \cite{Dong}.
Yamada et al. developed artificial skin surface ridges for detecting incipient slip by mimicking human tactile information from only several kinds of mechanoreceptors \cite{Yamada2001}.

Syntouch's BioTac is a biomimetic tactile sensor with an artificial skin filled with conductive fluid \cite{biotac}.
It has human-like vibrotactile and distributed sensation.
Using BioTac's elasticity and multimodality, Veiga et al. validated slip detection and grip stabilization \cite{Veiga,Veiga2}.
According to Su et al., BioTac's AC pressure sensor (an internal single hydrophonic vibration sensor \cite{microvibration}) detects incipient slip by locating the vibrotactile signal resulting from incipient slip \cite{Su}.

Although vibrotactile signal is easy to detect even with a single vibration sensor, it is susceptible to noise because it can only be observed after the robot's action or an external force is applied (\reffig{fig: comparison}, top figure).
Furthermore, the vibrotactile signal is less likely to appear when the friction coefficient and the gripping force are low.
Distinguishing between incipient and gross slip is challenging when a vibrotactile signal is detected \cite{review_slip}.
As a result, the grip control's performance is often evaluated using external measurement devices or comparing the results with numerical simulations based on the outcomes of the incipient slip detection \cite{review_slip}.

\subsubsection{Pressure distribution}
Narita et al. constructed a mechanical model of grip control based on the stick ratio, enabling control of the incipient slip in the rotational direction \cite{sony}.
Ikeda et al. demonstrated that a vision-based tactile sensor can measure the slip margin through the eccentricity of the contact surface and control gripping \cite{Ikeda}.
With a sim2real method, Griffa et al. pre-correlated the information from tactile sensor images with stress distribution, a strategy that enabled them to calculate the stick ratio in a model-based way \cite{leverage}.

\begin{figure}
  \centering
  \includegraphics[width=0.8\hsize]{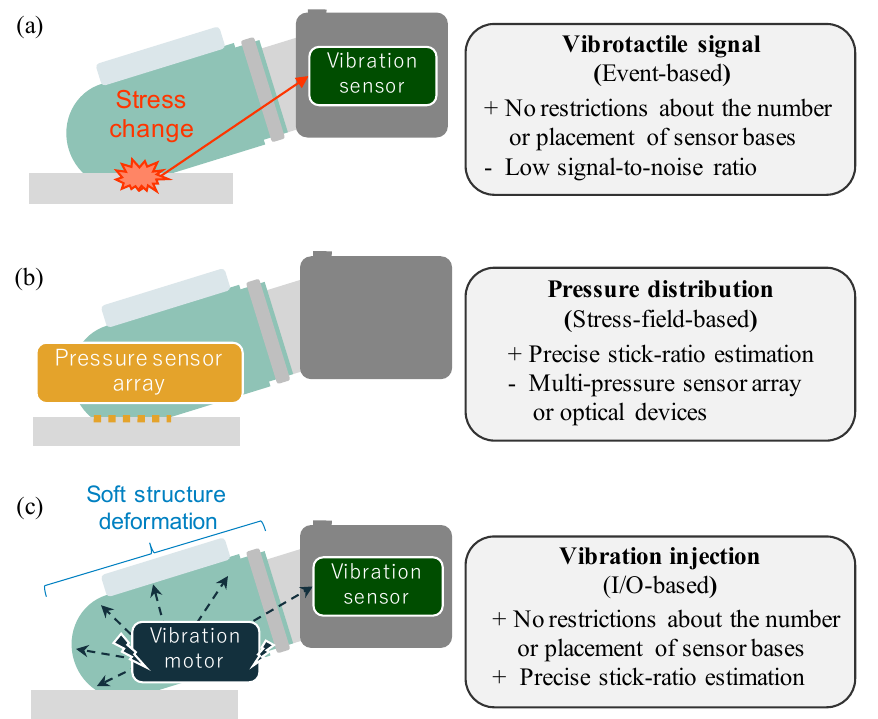}
  \vspace{-2mm}
  \caption{Comparison of incipient slip detection methods: (a)(b) Each conventional method has its advantages and disadvantages. (c) Proposed method has a mechanism that incorporates the best of both.}
  \label{fig: comparison}
\end{figure}

Informative methods that utilize stick ratio and dynamics models can be applied to the feature of pressure distribution, depending on the force sensor arrays or optical vision-based tactile information (\reffig{fig: comparison}, middle figure).
However, such sensing systems incur measurement-side costs, including large optical systems and the high-density integration of force sensor arrays \cite{Shimonomura}.
Moreover, the applicability of model-based methods is limited since they require that an object's geometry and the sensors' material properties be predetermined.

\subsection{Propagated vibration for tactile perception}
Ando et al. proposed a method for measuring stress based on the changes in resonance frequency \cite{ARTC}.
Their method employs an acoustic tensor cell tactile sensor that reverberates ultrasonic waves inside a cavity.
Mikogai et al. proposed a method for estimating the contact point of a soft robot based on frequency spectrum changes \cite{softrobot}. 
It uses the noise caused by internal air pressure as a vibration source.
Z{\"o}ller et al. proposed a method that identifies a pneumatic soft robot's contact position, contact force, and material.
Their method uses a microphone mounted inside the robot \cite{softfinger}.
Komeno and Matsubara proposed texture identification based on the changes of the characteristics of the propagated vibration through a tactile sensor's soft structure \cite{Komeno}.
Vibrations were applied to a cantilever beam using an eccentric motor and detected with an accelerometer to measure the triaxial stresses in the beam, as described by Kuang et al. \cite{3-axis}.

These studies utilized propagated vibration through a soft structure as a medium and measured the changes in the frequency-domain features using a single microphone or accelerometer or sonic pressure sensors.
This approach, which enables the extraction of such external interaction information as contact and stress, significantly reduces the number of sensor bodies by increasing the temporal resolution.
The advantage is similar to vibrotactile sensor systems.

\red{
Similar to the present study, Liu et al. proposed a smart skin that detects the incipient slip of a contact surface by applying ultrasonic waves to a beam structure attached to a soft object \cite{active}.
However, their slip detection experiments controlled the normal force applied to the contact surface to a specific value.
This requirement may be because the normal force suppresses the vibration propagation characteristics of the beam structure.
However, in practical use, controlling the normal force to a desired value for incipient slip detection is a challenge.
On the other hand, by using a biomimetic structure, our method alleviates the problem of suppression of vibration propagation by the normal forces (\reffig{fig: comparison} bottom figure).
As a result, our method is able to detect incipient slip despite dynamic fluctuations in normal force.
This will be verified in the experimental results presented later.
}

\section{Proposed system}
\subsection{Mechanism}
\begin{figure}
  \centering
  \includegraphics[width=0.9\hsize]{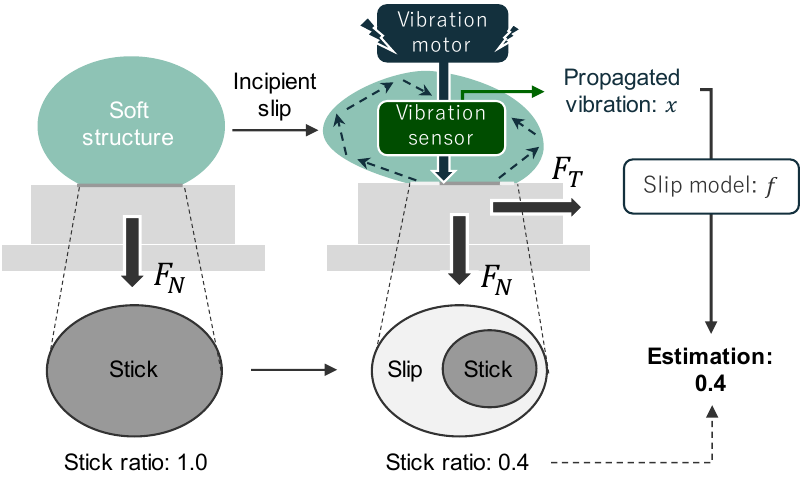}
  \vspace{-2mm}
  \caption{Mechanism of our method for incipient slip detection using vibration injection: Completely stick contact surface without tangential force $F_T$ changes into an incipient slip with $F_T$. Estimating stick ratio with propagated vibration enables us to detect incipient slip.}
  \label{fig: mechanism}
\end{figure}

Our method is based on the macroscopic deformation of soft tactile sensors due to incipient slip.
As shown in \reffig{fig: mechanism}, when normal force $F_N$ is constant, an increase in tangential force $F_T$ causes incipient slip and decreases the stick ratio.
A soft tactile sensor is subjected to variable stress based on the stick ratio, leading to the deformation of the entire soft structure through the contact surface.
Therefore, the stick ratio can be indirectly estimated from the deformation quantity of the soft structure.
Specifically, deformation measurement refers to the features of the vibration medium utilized in soft robotics \cite{softrobot,softfinger,Komeno}.
A single vibration sensor can measure the vibration propagation within a soft structure, and the frequency response depends on the current deformation state.
By combining the frequency spectrum with stick ratio's information and developing a regression model, the system can accurately detect incipient slips.

\subsection{System implementation}
The proposed method requires the following components in an actual machine:
\begin{itemize}
    \item A soft structure that macroscopically reflects the stick ratio's changes due to incipient slip.
    \item A vibration generator with a wide frequency response.
    \item A vibration sensor with high temporal resolution.
\end{itemize}

To achieve a surface contact, a tactile sensor's surface is typically composed of elastomer material.
Despite being flat, its small volume and high elastic modulus cause difficulties in deformation with limited frequency response.
It is desirable to have a tactile sensor with a soft structure characterized by a low elastic modulus, low viscosity, and adequate size to magnify the stick ratio's changes due to microscopic incipient slip.
Furthermore, it is imperative to have an actuator that injects mechanical vibration into the soft structure and a vibration sensor with high temporal resolution.
From the system identification perspective, such white signals and M-sequence signals are promising candidates for applied vibrations.
In contrast, since coils and eccentric motors can only be stimulated in the neighborhood of their resonant frequency, a specific actuator with a good frequency response is needed as a vibration generator.

A biomimetic tactile sensor (Syntouch, BioTac) and a small PZT motor (Cedrat Technology, APA50XS) were used as hardware to achieve these components.
A BioTac is composed of a soft structure that consists of artificial skin and conductive fluid and exhibits excellent deformation and vibrotactile signal propagation characteristics \cite{biotac,Su,Veiga,Veiga2}.
In contrast to ordinary coil or eccentric vibration motors, a PZT motor can operate at frequencies other than the resonant frequency and generate vibrations with arbitrary frequency components \cite{Kurita,Komeno}.
\red{
We utilize Gaussian noise as a vibration signal within 10 to 1100 Hz, which satisfies all sampling frequencies of BioTac.
}
\reffig{fig: exp_slip} shows that the PZT motor is mounted on the side of the BioTac to propagate vibration to its soft structure.
We adopted a frequency spectrum as the propagation characteristics of the applied vibration, based on research in soft robots \cite{softrobot,softfinger,Komeno}.

\section{Experiment}
We evaluated our proposed method's stick-ratio estimation and stick-ratio stabilization performance and also compared its performance to conventional methods by employing vibrotactile signal and pressure distribution techniques.

\subsection{Setting}
\begin{figure}
  \centering
  \includegraphics[width=0.9\hsize]{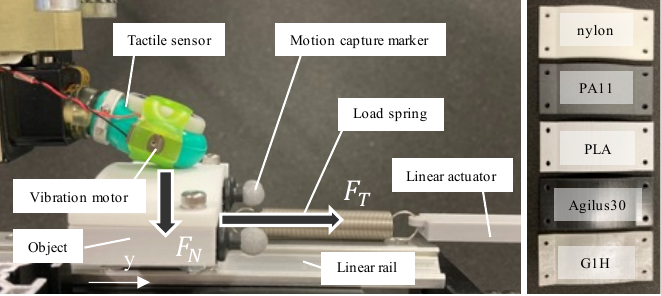}
  \vspace{-2mm}
  \caption{Slip experimental setup: Linear actuator and spring make tangential force $F_T$ and motion capture system detects precise object position $y$. Objects have identical geometry but are made from different materials by 3D printers for various friction coefficients and softness.}
  \label{fig: exp_slip}
\end{figure}

The experimental setup is shown in \reffig{fig: exp_slip}.
The proposed system consists of a robotic arm (Universal Robots, UR5) that stabilizes an object by BioTac, which measures the tactile information \textbf{x}.
Although BioTac has multiple sensor modalities \cite{biotac,microvibration}, in this study, AC pressure $P_{\mathrm{AC}}\in\mathbb{R}$, DC pressure $P_{\mathrm{DC}}\in\mathbb{R}$, and electrodes $E\in\mathbb{R}^{19}$ were used as slip-related information.
The BioTac moves vertically, and normal force $F_N$ is calculated based on the value of DC pressure $P_{\mathrm{DC}}$.

An object is attached to a linear rail and can only move horizontally.
A motion capture system measures the object's position $y$ (OptiTrack, Flex13).
The object's tip is linked to a linear actuator that moves horizontally by a tension spring.
As a result, this creates external tangential force $F_T$ that can change the stick ratio of the contact surface.

The object is comprised of the five materials shown in \reffig{fig: exp_slip}.
All have convex shapes with a 150 mm curvature made by a 3D printer.
Each material has different friction or elasticity.
This difference checks whether the vibration injection depends on the object's physical properties to impair its propagation ability.
In other words, it verifies whether the propagation of the vibrations is robust to the object's viscoelasticity or its surface roughness.

A series of experiments discretely defined the actions of the robot and the linear actuators.

\subsection{Stick-ratio estimation}
This study estimates stick ratio $s_t$ from tactile information $\textbf{x}_t$ at time $t$.
Its estimation is expressed by the following equation, assuming slip model $f$ and a feature function of tactile information, $\phi$:
\begin{equation}
    s_t = f(\phi(\textbf{x}_{t-\mathrm{T}:t})).
    \label{eqn: slip_pred}
\end{equation}
$\mathrm{T}$ refers to the sampling window size.
stick ratio $s_t$ is identified in the dataset based on tangential force $F_T$ and position $y$ of the object (\reffig{fig: slip_time}).
Further details regarding this setting is provided in later sections.

\subsubsection{Data collection}
During the experiment, the robot applies normal force $F_N$ to stabilize an object, and tangential force $F_T$ is subsequently exerted on it by a linear actuator.
The linear actuator is moved a fixed distance, discretely and monotonically increasing $F_T$ the maximum movement distance (23 mm, 450 steps).
During the sampling period, we measured tactile information $\textbf{x}$ of the BioTac, the position of object $y$, and $F_T$, which is determined by the relative position of the linear actuator and the object.
\red{
We collected 100 samples of each material, and sampling window $\mathrm{T}$ is 1.0 sec.
}
The experiment uses Gaussian noise $\varepsilon$ for the vibration, and the PZT motor sets the amplitude to its maximum rating.
The standard normal distribution with intensity $I=0$ dB is the vibration's sampling source:
\begin{equation}
  \varepsilon \sim 10^{I/20}\mathcal{N}(0,1).
  \label{eq: noise}
\end{equation}
We collected for comparison two patterns of data: one with a vibration injection and one without.

\begin{figure}
  \centering
  \includegraphics[width=0.9\hsize]{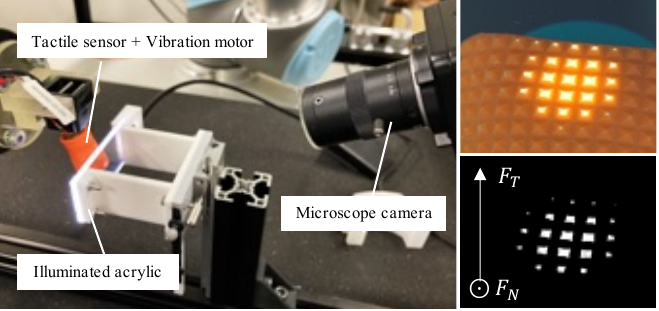}
  \vspace{-2mm}
  \caption{Contact surface observation based on optical tactile sensor mechanism}
  \label{fig: exp_camera}
\end{figure}

\begin{figure}
  \centering
  \includegraphics[width=0.6\hsize]{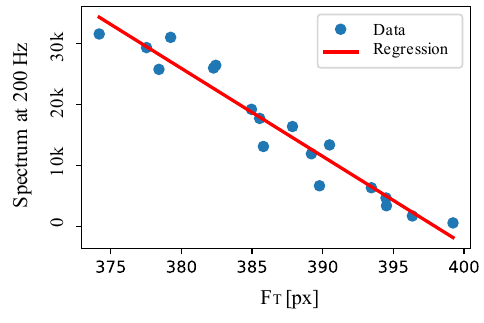}
  \vspace{-2mm}
  \caption{Linear correlation for tangential force $F_T$ and frequency spectrum at 200 Hz: $R=0.97$}
  \label{fig: camera_spec}
\end{figure}

\subsubsection{Labeling \redbf{pseudo-stick ratio}}
The proposed method detects incipient slips based on the \bluebf{stick ratio}.
However, measuring its precise value is challenging and requires identification through a contact model or optical observation \cite{Ikeda,leverage,sony}.
Since BioTac hasn't yet proposed a method to determine these values, this study defines the \redbf{pseudo-stick ratio} as a ground truth of \bluebf{stick ratio}.
Assuming that normal force $F_N$ and tangential force $F_T$ for the soft structure are sufficient features to calculate the \bluebf{stick ratio} during incipient slip, we defined a completely stick contact surface $F_T=0$ as $s=1.0$, and gross slip when object position $y$ exceeds the threshold value and tangential force $F_T^{slip}$ is given as $s=0.0$.
The \redbf{pseudo-stick ratio}'s transition through the incipient slip is linearly interpolated between the complete stick and the gross slip based on tangential force $F_T$:
\begin{equation}
    s = 1-\frac{F_T}{F_T^{slip}}.
    \label{eqn: SR}
\end{equation}
\redbf{pseudo-stick ratio} $s$ is used to label the ground truth of the whole data.
The validity of this linear completion is supported by the experimental results.
As illustrated in \reffig{fig: exp_camera}, the contact surface was observed directly using optical observation to study how a tactile sensor with soft structure deformation changes its slip state.
With a transparent acrylic plate and LEDs, along with a microscope camera (HOZAN, L-837), we observed the movement of the contact surface that corresponded to the \bluebf{stick ratio}.
The resulting optical information provides the coordinates and intensity, and they were measured as normal and tangential forces \cite{camera,Shimonomura}.
$F_N$ was used as the light intensity, and $F_T$ was used as the displacement of the center of gravity on the horizontal axis.
A sawtooth wave with a discrete frequency spectrum was utilized as a vibration injection signal.
The frequency corresponding to $F_T$ was extracted and analyzed at 100 Hz intervals.

\begin{figure}
  \centering
  \includegraphics[width=0.9\hsize]{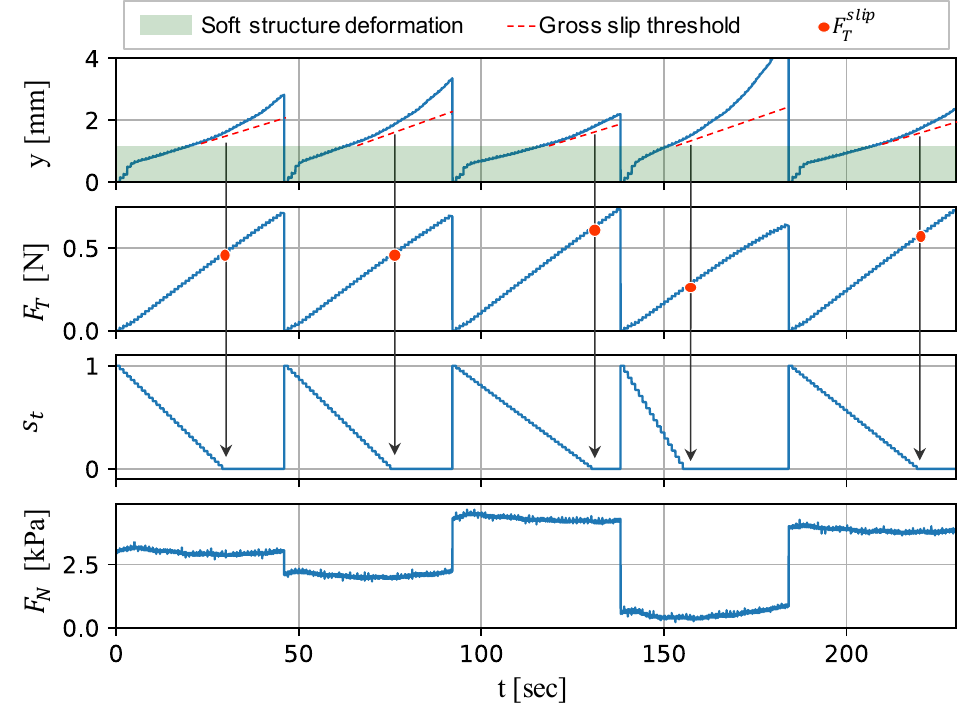}
  \vspace{-2mm}
  \caption{Time series of object position $y$, tangential force $F_T$, \redbf{pseudo-stick ratio} $s$, and normal force $F_N$ during data collection: Red dotted lines show threshold of gross slip for $y$ derivative, which defines $F_T^{slip}$ and $s$ = 0.0. Thus, \refeqn{eqn: SR} can define the entire \redbf{pseudo-stick ratio}.}
  \label{fig: slip_time}
\end{figure}

\reffig{fig: camera_spec} shows the correspondence between $F_T$ calculated from the optical information of the contact surface and the spectrum at 200 Hz, the most sensitive frequency band in BioTac.
As $F_T$ increases, its spectrum decreases.
We also confirmed that they are linearly correlated with a coefficient of determination of $R^2=0.97$.
This result indicates that the \bluebf{stick ratio} and the frequency spectral change can be represented as linear features, as in \eqref{eqn: SR}.

\reffig{fig: slip_time} illustrates travel distance $y$, tangential force $F_T$, \redbf{pseudo-stick ratio} $s$, and normal force $F_N$ for the five sample series.
The tension spring discretely and monotonically increases $F_T$ every $\mathrm{T}$ seconds, decreasing $s$.
\red{
Since the tactile sensor in this study has a soft structure, the coordinates of the object change due to its deformation even when gross slip is not occurring (upper row, green area).
Therefore, the amount of movement of 0.02 mm / 500 msec in the second half of the sampling time $\mathrm{T}$ is defined as gross slip (upper row, red dotted lines).
This behavior was determined by trial and error, taking into account the measurement error of the motion capture system, as a behavior that can be clearly defined as gross slip in all sample series.
}
This definition also determines the point where the tangential force is $F_T^{slip}$ (second row, red circles).
The \redbf{pseudo-stick ratio} can be labeled using \refeqn{eqn: SR} in conjunction with the fact that $s=1.0$ when $t=0.0$ (third row).
\red{
While the linear actuator moves a constant distance for each sampling period $\mathrm{T}$, $F_N$ is uniformly randomized from discretized candidates at intervals of 0.4 kPa within the range from 1.1 to 5.9 kPa and controlled by the end-effector's position (bottom) to cover the state-action space in the stabilization control explained below.
Therefore, the condition of the contact surface varies with $F_N$, and the amount of object movement also varies because the frictional force generated is not constant, resulting in variations in the way $F_T$ is given (second row).
}

\subsubsection{Learning the slip models}
Slip model $f$ based on the \redbf{pseudo-stick ratio} $s$ is derived from the collected data.
Feature function $\phi$ computes the fast Fourier transform frequency spectrum for vibrotactile signal within 10-1100 Hz or a time-averaged process for pressure distribution.
The $s$ is defined discretely within a sampling interval of $T=500$ msec.
The Support Vector Regression (SVR), which is utilized as $f$.
We determined the optimal kernel and the parameters for each dataset by a grid search.

\subsubsection{Evaluation}
The proposed and comparison methods use various BioTac sensor modalities to obtain tactile information $\textbf{x}$.
The experiment employed the following comparison methods, which utilize :
\begin{itemize}
    \item vibrotactile signal ($P_{AC}$),
    \item pressure distribution ($E_{19,10,4,1}$),
    \item our proposed method ($P_{AC}$ with vibration injection).
\end{itemize}
Our proposed method used AC pressure $P_{AC}$ as a sensor to propagate vibrations.
In contrast, the comparison method used $P_{AC}$ without a vibration as a vibrotactile sensor and captured the pressure distribution using 19 electrodes $E_{19}\in\mathbb{R}^{19}$.
Furthermore, our study compared four distinct pressure distribution E sensors with reduced resolution.
The data were sparsified by removing them from all of the E electrodes at equal intervals: 50\% for $E_{10}\in\mathbb{R}^{10}$, 20\% for $E_4\in\mathbb{R}^{4}$, and single electrode $E_1\in\mathbb{R}^{1}$ in the center.
This procedure verifies the proposed method's performance and spatial resolution accuracy when dealing with pressure distribution.

For each sample series, the Root Mean Square Error (RMSE) for the estimation error was evaluated using 10\% of the dataset as test data.
Since huge estimation errors can destabilize the gripping object and cause irreversible damage, we also compared and tested the top 10\% of RMSE, a value that represents the worst case for the estimation error.
Additionally, we compared the statistical significance of the proposed method with the others using a t-test.
The RMSE was calculated for each sample series, including all the materials, resulting from 10\% of the 500 samples.

\begin{figure*}
  \centering
  \includegraphics[width=0.9\hsize]{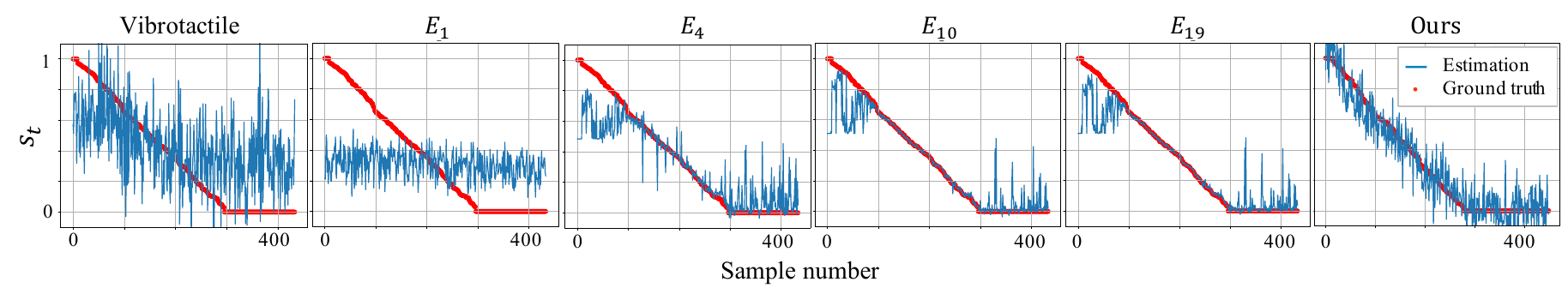}
  \vspace{-2mm}
  \caption{Results for stick-ratio estimation in each method. \red{The horizontal axis is the sample index when sorted in order of \bluebf{stick ratio}.}}
  \label{fig: reg}
\end{figure*}

\begin{figure}
  \centering
  \includegraphics[width=0.8\hsize]{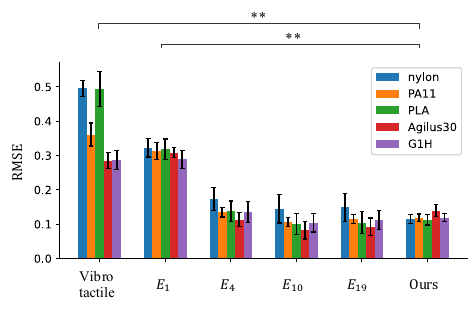}
  \vspace{-2mm}
  \caption{RMSE of stick-ratio estimation in each method for all materials (** means $p < 0.01$).}
  \label{fig: rmse_all}
\end{figure}

\subsection{Stick-ratio stabilization}
We experimentally applied slip model $f$ to actual stick-ratio stabilization control.
In this experiment, the feedback controller determined normal force $F_N$ that maintains target \bluebf{stick ratio} $s_d$ based on $s$ estimated in real-time.
This expectation aims to prevent gross object slip and maintain minimum required $F_N$.
The experiment verifies the effectiveness of the slip estimation and its robustness to the changes in the contact region following the grip control.

Tangential force $F_T$ is applied as in the data collection.
Controlling action $a(s_t)$, in contrast, relies on $s_t$ estimated by vibration injection:
\begin{equation}
    a(s_t) = k(s_t - s_d),
    \label{eqn: action}
\end{equation}
where $k$ is the gripping force gain and the value converging to $s_d$ is determined manually \cite{Ikeda}.
The stabilization continues until $F_N$ reaches a safety limit.

\subsubsection{Evaluation}
We evaluated object travel distance $y$ and normal force $F_N$ of 10 samples to determine the stick-ratio stabilization results.
\red{
Based on the average distance moved by the object in the dataset when fitting the definition of gross slip ($1.5\pm0.4$ mm) and the maximum normal force $F_N$ (5.9 kPa), we defined failure as when y $>$ 1.5 mm and $F_N >$ 6.0 kPa.
}

Using the degree of achievement of each stabilization control as a score, we calculated the achievement score with the evaluation function:
\begin{equation}
    \mathrm{score} = \left( w_1 + \mathrm{success rate} \right) * \left( \frac{w_2}{y} + \frac{w_3}{F_N} \right),
    \label{eqn: reward}
\end{equation}
\red{
where $w_1$ is a parameter makes score positive and $w_2,w_3$ are scaling parameters to match the order of $y$ and $F_N$.
}
This function calculates a unified scalar value that treats the three parameters equivalently in minimizing $y$ and $F_N$ and maximizing the success rate.
Even if a gross slip is suppressed, a penalty is imposed when the pressure is high, and vice versa.
The score is also evaluated by examining the significant differences between the proposed and others.

To examine the worst-case scenario of the gross slip, we also conducted an additional control method using no action ($a=0$) where no stabilization controls were performed.

\section{Results}
\subsection{Stick-ratio estimation}
\reffig{fig: reg} displays a graph of the ground truth values and the estimated values of the $s$.
\reffig{fig: rmse_all} shows the RMSE, which indicates the estimation error for each method and material.
Vibrotactile signal and $E_{1}$ have a large variance, indicating that no estimation was achieved.
On the other hand, our proposed method, $E_{19},E_{10}$, and $E_{4}$ successfully estimated the $s$ with low RMSE, and no significant difference was identified among them.
Although the proposed method has stable estimation with minor errors on average over the entire range, huge errors are seen above $s=0.7$ for $E_{19}, E_{10}$, and $E_{4}$.
\red{
These results suggests that in conventional methods, there are cases where the sensor's transducer cannot capture the microscopic slip state of the contact surface.
Vibration injection can detect slips in any situation because it extracts macroscopic deformation of the entire soft structure as a medium, including the protective cover.
}

\begin{figure}
  \centering
  \includegraphics[width=0.8\hsize]{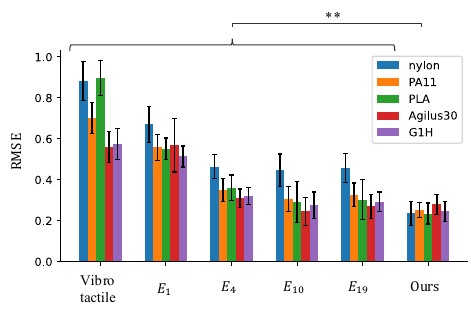}
  \vspace{-2mm}
  \caption{RMSE in top 10\% of stick-ratio estimation in each method for all materials (** means $p < 0.01$)}
  \label{fig: rmse_percentile}
\end{figure}

\begin{figure}
 \centering
 \includegraphics[width=0.8\hsize]{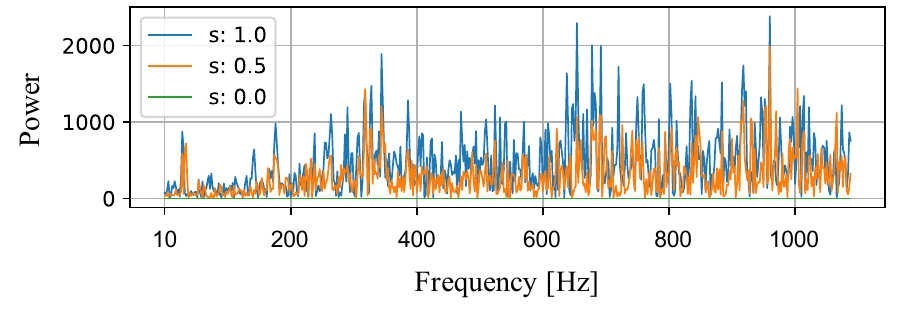}
 \vspace{-2mm}
 \caption{\red{The injected vibration feature for estimation in frequency domain}}
 \label{fig: freq}
\end{figure}

\reffig{fig: rmse_percentile} shows the top 10\% of RMSE, representing the worst case for the estimation error.
The proposed method records the lowest RMSE, which is significantly lower than that of the pressure distribution.
This difference is expected to affect subsequent stick-ratio stabilization control.

The five materials in \reffig{fig: exp_slip} exhibited no drastic differences or trends.
This result shows that vibration injection is a robust method that can be applied regardless of the hardness and surface properties of an object.

\red{
Fig. \ref{fig: freq} shows the average frequency spectrum (based on $s = 0.0$) of 100 samples in proposed method.
It can be seen that the intensity of the spectrum changes with $s$.
}

In summary, our method's incipient slip detection performance resembles that of $E_{19}, E_{10}$, and $E_{4}$, although it outperforms them in terms of extreme estimation error.

\subsection{Stick-ratio stabilization}
The stick-ratio stabilization results are shown in \reftab{tbl: stab}.
\red{
The arrows indicate which of the larger or smaller values contributes to the performance of the control.
The step is the number of execution steps of the task and indicates how long the experiment continues without failure (MAX: 450).
The $y$ and $F_N$ are the average values in the last sampling step.
}

\red{
For a fair performance comparison, we set a target value of $s_d=0.3$, where the estimation error for both methods is low.
While the estimation error of the proposed method has a similar amount over the entire region, a significantly larger error occurs around $s_d=0.7$ for the pressure distribution $E_{19,10}$.
Therefore, similar results are expected for any value below $s_d=0.7$, but performance deterioration is expected for the compared methods above this level.
}
The proposed method, $E_{19}$, and $E_{10}$ had a high success rate, and position $y$, which was caused by the object slip, was kept around 1.4 mm, a value within the deformable range of the soft structure itself.
Such successful slip control is indicated when objects move about 2.5 mm in the case of no action where the object was not controlled.
In contrast, stabilization control failed most with $E_{4}, E_{1}$, and vibrotactile signal.
Action $a$ in vibrotactile signal and $E_{1}$ led to gripping pressure $F_N$ that exceeded the safety limit at approximately 6.0 kPa and stopped.
In contrast, a stabilization action was barely performed in $E_{4}$, leading to a failure of stabilization.

\begin{table}
 \centering
 \caption{Stabilization achievement}
 \label{tbl: stab}
 \vspace{-4mm}
 \begin{tabular}[t]{c|>{\centering}p{3em}>{\centering}p{3em}ccccc}
  \hline
	Method                     &No action &Vibro tactile &$E_1$ &$E_{4}$ &$E_{10}$ &$E_{19}$ &Ours\\\hline
        Succ. rate $\uparrow$    &- &0.3 &0.0 &0.1 &1.0 &0.8 &1.0\\
        y [mm]       $\downarrow$  &2.48 &1.32 &0.89 &2.52 &1.55 &1.40 &1.38\\
        $F_N$ [kPa]  $\downarrow$  &2.94 &5.96 &6.47 &3.59 &5.23 &6.24 &4.50\\
        Step         $\uparrow$    &450 &397 &139 &450 &450 &435 &450\\\hline
 \end{tabular}
\end{table}

\begin{figure}
  \centering
  \includegraphics[width=0.8\hsize]{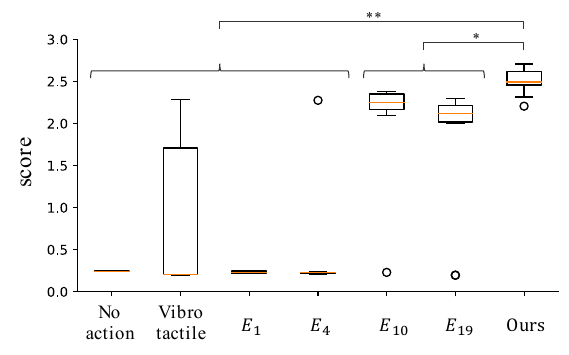}
  \vspace{-2mm}
  \caption{Score of stick-ratio stabilization in each method (** means $p < 0.01$ and * means $p < 0.05$).}
  \label{fig: reward}
\end{figure}

\begin{figure}
  \centering
  \includegraphics[width=0.9\hsize]{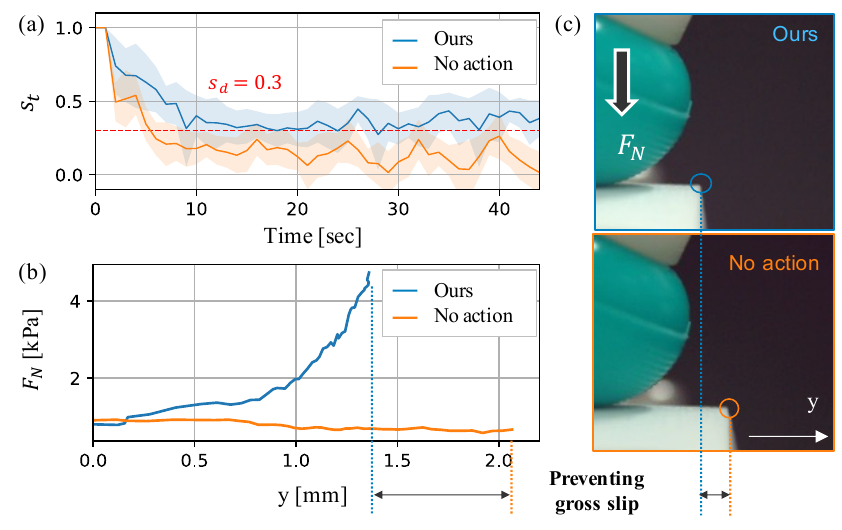}
  \vspace{-2mm}
  \caption{Results of stick-ratio stabilization: (a) Estimated \bluebf{stick ratio} $s$, (b) transition of stabilization action $F_N$ and object movement $y$, and (c) pictures of actual object movement.}
  \label{fig: snap}
\end{figure}

\reffig{fig: reward} shows a score comparison with parameters $\{w_1,w_2,w_3\} = \{0.1,10,1\}$.
The proposed method is significantly different from $E_{19}$ and $E_{10}$ at $p < 0.05$ and other methods at $p < 0.01$.
This trend corresponds to the RMSE analysis of the top 10\% of the estimation errors shown in \reffig{fig: rmse_percentile}, 
suggesting that although the average incipient slip detection performances are similar, methods with extreme errors compromise the control performance.
\red{
Here, we confirmed that changing the absolute value of the score and the relative scale between $y$ and $F_N$ by increasing or decreasing the parameters $w_1,w_2,w_3$ in the positive range does not change the trend of the graph.
}

\reffig{fig: snap} shows the control transition and the final state of the proposed method (blue) and no action (orange).
The upper figure is the estimated \bluebf{stick ratio}'s values, and the lower is object movement $y$ vs. normal force $F_N$.
The pictures on the right show considerable deviation caused by gross slip.
The proposed method gradually regulates $F_N$ to approach the target $s_d=0.3$. 
This result indicates that the incipient slip detection is accurately functioning.
Compared to no action, $y$ is suppressed.
Therefore, the proposed method successfully prevents gross slip and achieves stabilization control.
\red{
$F_N$ changes dynamically during griping control, so $F_T$ has a different time-series change than during data collection, but convergence to the target value has been confirmed.
It can be said that changes in $F_T$ do not much affect the estimation within the range of what has been learned.
}

In \reffig{fig: rmse_all}, \ref{fig: rmse_percentile}, and \ref{fig: reward}, as the resolution appears to decrease, pressure distribution $E_{19,10,4,1}$ detects less incipient slip and has a lowered control performance.
The proposed method's best performance almost matches $E_{19}$, comparable to its highly distributed tactile resolution.

The proposed method demonstrates effective stabilization control based on real-time incipient slip detection and significantly outperforms the pressure distribution.
This result reflects that the proposed method keeps the estimation error small over the entire stick-ratio range.

\section{Discussions}

\red{
In this study, the soft structure is subjected to high stress by varying the tangential and normal forces.
Experiments were conducted based on the assumption that the performance is maximized near the maximum rating of the PZT motor, referring to the slit width classification task in the study \cite{Komeno} using a system similar to that used in this study.
It is considered necessary to determine the optimal intensity for tasks with small stress variation, such as the texture classification task in \cite{Komeno}, depending on the vibration motor's rated capacity.
On the other hand, the frequency of the injected vibration was not considered an issue to adjust, as the white noise could be used to vibrate the entire sampling range of the sensor to handle various situations. 
}

\red{
The vibration motor, which must be provided separately from the vibration sensor, incurs extra costs.
However, they are not exposed to mechanical loads and can be easily mounted on the side since they are not transducers.
Conversely, pressure distribution sensors require a pre-embedded sensor array, which increases the risk of damage, mainly when located very close to the contact surface, as with BioTac's electrodes.
Therefore, the vibration injection method has an advantage in operating costs, although the initial cost is expensive.
}

\red{
The experiments are based on preliminary characterization and data-driven design for the Biotac sensor.
We expect that a similar process flow will be valid for use of other tactile sensors than Biotac, although some adjustments in parameters and settings may be necessary.
For example, for regression in labeling \redbf{pseudo-stick ratio}, a more complex functional model such as a polynomial or exponential function may be appropriate instead of a linear function, depending on different sensor materials/constructions.
}

\red{
To detect informative propagating vibrations, a medium with good frequency response and a high temporal resolution sensor are required.
However, there are no commercial or open-source tactile sensors other than BioTac, meets these requirements with its biomimetic structure and internal vibration sensor.
For example, the elastomer material of the GelSight \cite{Dong} has a poor frequency response due to its high elasticity, and TacTip \cite{tactip} makes it difficult to optically observe the displacement of the vibration regarding camera resolution and sampling frequency.
In the future, it is expected that a specific tactile sensor suitable for vibration injection applications will be manufactured, and its general properties will be confirmed to be widely used as a new platform.
}

\section{Conclusion}
We proposed a new incipient slip detection method using vibration injection for a soft structure deformation corresponding to the \bluebf{stick ratio} with a single vibration sensor.
In the actual system, slip models were learned from the propagated vibration characteristics.
In validation experiments for stick-ratio estimation and stabilization control, we confirmed that the proposed method achieved the same performance and, in some cases, even outperformed the conventional methods.
We expect our proposed method to enable incipient slip detection and stick-ratio stabilization control in any environment without a high spatial resolution sensor.

\bibliographystyle{IEEEtran}
\bibliography{IEEEabrv,mybibfile}

\begin{thebibliography}{10}
\providecommand{\url}[1]{#1}
\csname url@rmstyle\endcsname
\providecommand{\newblock}{\relax}
\providecommand{\bibinfo}[2]{#2}
\providecommand\BIBentrySTDinterwordspacing{\spaceskip=0pt\relax}
\providecommand\BIBentryALTinterwordstretchfactor{4}
\providecommand\BIBentryALTinterwordspacing{\spaceskip=\fontdimen2\font plus
\BIBentryALTinterwordstretchfactor\fontdimen3\font minus \fontdimen4\font\relax}
\providecommand\BIBforeignlanguage[2]{{%
\expandafter\ifx\csname l@#1\endcsname\relax
\typeout{** WARNING: IEEEtran.bst: No hyphenation pattern has been}%
\typeout{** loaded for the language `#1'. Using the pattern for}%
\typeout{** the default language instead.}%
\else
\language=\csname l@#1\endcsname
\fi
#2}}

\bibitem{Johansson}
R.~S. Johansson and G.~Westling, ``\BIBforeignlanguage{en}{Roles of glabrous skin receptors and sensorimotor memory in automatic control of precision grip when lifting rougher or more slippery objects},'' \emph{\BIBforeignlanguage{en}{Experimental brain research. Experimentelle Hirnforschung. Experimentation cerebrale}}, vol.~56, no.~3, pp. 550--564, 1984.

\bibitem{review_slip}
W.~Chen, H.~Khamis, I.~Birznieks, N.~F. Lepora, and S.~J. Redmond, ``Tactile sensors for friction estimation and incipient slip detection---toward dexterous robotic manipulation: A review,'' \emph{IEEE sensors journal}, vol.~18, no.~22, pp. 9049--9064, Nov. 2018.

\bibitem{review_dexterous}
Z.~Kappassov, J.-A. Corrales, and V.~Perdereau, ``Tactile sensing in dexterous robot hands --- review,'' \emph{Robotics and autonomous systems}, vol.~74, pp. 195--220, Dec. 2015.

\bibitem{Dong}
S.~Dong, W.~Yuan, and E.~H. Adelson, ``Improved gelsight tactile sensor for measuring geometry and slip,'' in \emph{2017 IEEE/RSJ International Conference on Intelligent Robots and Systems}, Sept. 2017, pp. 137--144.

\bibitem{Yamada2001}
Y.~Yamada, I.~Fujimoto, T.~Morizono, Y.~Umetani, T.~Maeno, and D.~Yamada, ``Development of artificial skin surface ridges with vibrotactile sensing elements for incipient slip detection,'' in \emph{Conference Documentation International Conference on Multisensor Fusion and Integration for Intelligent Systems}, Aug. 2001, pp. 251--257.

\bibitem{Su}
Z.~Su, K.~Hausman, Y.~Chebotar, A.~Molchanov, G.~E. Loeb, G.~S. Sukhatme, and S.~Schaal, ``Force estimation and slip detection/classification for grip control using a biomimetic tactile sensor,'' in \emph{2015 IEEE-RAS 15th International Conference on Humanoid Robots}, Nov. 2015, pp. 297--303.

\bibitem{microvibration}
J.~A. Fishel, V.~J. Santos, and G.~E. Loeb, ``A robust micro-vibration sensor for biomimetic fingertips,'' in \emph{2008 2nd IEEE RAS EMBS International Conference on Biomedical Robotics and Biomechatronics}, Oct. 2008, pp. 659--663.

\bibitem{sony}
T.~Narita, S.~Nagakari, W.~Conus, T.~Tsuboi, and K.~Nagasaka, ``Theoretical derivation and realization of adaptive grasping based on rotational incipient slip detection,'' in \emph{2020 IEEE International Conference on Robotics and Automation}, May 2020, pp. 531--537.

\bibitem{Ikeda}
A.~Ikeda, Y.~Kurita, J.~Ueda, Y.~Matsumoto, and T.~Ogasawara, ``Grip force control for an elastic finger using vision-based incipient slip feedback,'' in \emph{2004 IEEE/RSJ International Conference on Intelligent Robots and Systems}, vol.~1, Sept. 2004, pp. 810--815.

\bibitem{leverage}
P.~Griffa, C.~Sferrazza, and R.~D'Andrea, ``Leveraging distributed contact force measurements for slip detection: a physics-based approach enabled by a data-driven tactile sensor,'' in \emph{2022 International Conference on Robotics and Automation}, May 2022, pp. 4826--4832.

\bibitem{active}
C.~Liu, T.~M. Huh, S.~X. Chen, L.~Lu, F.~Kopsaftopoulos, M.~R. Cutkosky, and F.-K. Chang, ``Design of active sensing smart skin for incipient slip detection in robotics applications,'' \emph{IEEE/ASME Transactions on Mechatronics}, vol.~28, no.~3, pp. 1766--1777, June 2023.

\bibitem{softrobot}
S.~Mikogai, B.~D.~C. Kazumi, and K.~Takemura, ``Contact point estimation along air tube based on acoustic sensing of pneumatic system noise,'' \emph{IEEE Robotics and Automation Letters}, vol.~5, no.~3, pp. 4618--4625, July 2020.

\bibitem{softfinger}
G.~Z{\"o}ller, V.~Wall, and O.~Brock, ``Acoustic sensing for soft pneumatic actuators,'' in \emph{2018 IEEE/RSJ International Conference on Intelligent Robots and Systems}, Oct. 2018, pp. 6986--6991.

\bibitem{biotac}
C.~H. Lin, T.~W. Erickson, J.~A. Fishel, N.~Wettels, and G.~E. Loeb, ``Signal processing and fabrication of a biomimetic tactile sensor array with thermal, force and microvibration modalities,'' in \emph{2009 IEEE International Conference on Robotics and Biomimetics}, Dec. 2009, pp. 129--134.

\bibitem{Veiga}
F.~Veiga, J.~Peters, and T.~Hermans, ``\BIBforeignlanguage{en}{Grip stabilization of novel objects using slip prediction},'' \emph{\BIBforeignlanguage{en}{IEEE transactions on haptics}}, vol.~11, no.~4, pp. 531--542, Oct. 2018.

\bibitem{Veiga2}
F.~Veiga, B.~Edin, and J.~Peters, ``\BIBforeignlanguage{en}{Grip stabilization through independent finger tactile feedback control},'' \emph{\BIBforeignlanguage{en}{Sensors}}, vol.~20, no.~6, Mar. 2020.

\bibitem{Shimonomura}
K.~Shimonomura, ``\BIBforeignlanguage{en}{Tactile image sensors employing camera: A review},'' \emph{\BIBforeignlanguage{en}{Sensors}}, vol.~19, no.~18, Sept. 2019.

\bibitem{ARTC}
S.~Ando, H.~Shinoda, A.~Yonenaga, and J.~Terao, ``\BIBforeignlanguage{en}{Ultrasonic six-axis deformation sensing},'' \emph{\BIBforeignlanguage{en}{IEEE transactions on ultrasonics, ferroelectrics, and frequency control}}, vol.~48, no.~4, pp. 1031--1045, July 2001.

\bibitem{Komeno}
N.~Komeno and T.~Matsubara, ``Tactile perception based on injected vibration in soft sensor,'' \emph{IEEE Robotics and Automation Letters}, vol.~6, no.~3, pp. 5365--5372, July 2021.

\bibitem{3-axis}
W.~Kuang, M.~Yip, and J.~Zhang, ``Vibration-based multi-axis force sensing: Design, characterization, and modeling,'' \emph{IEEE Robotics and Automation Letters}, vol.~5, no.~2, pp. 3082--3089, Apr. 2020.

\bibitem{Kurita}
Y.~Kurita, M.~Shinohara, and J.~Ueda, ``Wearable sensorimotor enhancer for fingertip based on stochastic resonance effect,'' \emph{IEEE Transactions on Human-Machine Systems}, vol.~43, no.~3, pp. 333--337, May 2013.

\bibitem{camera}
M.~Ohka, H.~Kobayashi, J.~Takata, and Y.~Mitsuya, ``An experimental optical three-axis tactile sensor featured with hemispherical surface,'' \emph{Journal of Advanced Mechanical Design, Systems, and Manufacturing}, vol.~2, no.~5, pp. 860--873, 2008.

\bibitem{tactip}
J.~W. James, N.~Pestell, and N.~F. Lepora, ``Slip detection with a biomimetic tactile sensor,'' \emph{IEEE Robotics and Automation Letters}, vol.~3, no.~4, pp. 3340--3346, Oct. 2018.

\end{thebibliography}

\end{document}